\documentclass{article}
\usepackage{ijcai22}

\usepackage{times}

\usepackage{soul}
\usepackage{url}
\usepackage[hidelinks]{hyperref}
\usepackage[utf8]{inputenc}
\usepackage[small]{caption}
\usepackage{graphicx}
\usepackage{amsmath}
\usepackage{booktabs}
\title{A Survey on Machine Learning Approaches for Modelling Intuitive Physics}

\author{
Jiafei Duan$^1$\footnote{Equal Contributions}\and
Arijit Dasgupta$^2$$^*$\and
Jason Fischer$^3$\And
Cheston Tan$^{1,4}$\\
\affiliations
$^1$Institute for Infocomm Research, A*STAR\\
$^2$Department of Mechanical Engineering, National University of Singapore\\
$^3$Department of Psychological and Brain Sciences, Johns Hopkins University\\
$^4$Centre for Frontier AI Research, A*STAR
\emails
\{duan\_jiafei, cheston-tan\}@i2r.a-star.edu.sg,
arijit.dasgupta@u.nus.edu,
jason.fischer@jhu.edu
}

\begin{document}

\maketitle

\begin{abstract}
Research in cognitive science has provided extensive evidence of human cognitive ability in performing physical reasoning of objects from noisy perceptual inputs. Such a cognitive ability is commonly known as intuitive physics. With advancements in deep learning, there is an increasing interest in building intelligent systems that are capable of performing physical reasoning from a given scene for the purpose of building better AI systems. As a result, many contemporary approaches in modelling intuitive physics for machine cognition have been inspired by literature from cognitive science. Despite the wide range of work in physical reasoning for machine cognition, there is a scarcity of reviews that organize and group these deep learning approaches. Especially at the intersection of intuitive physics and artificial intelligence, there is a need to make sense of the diverse range of ideas and approaches. Therefore, this paper presents a comprehensive survey of recent advances and techniques in intuitive physics-inspired deep learning approaches for physical reasoning. The survey will first categorize existing deep learning approaches into three facets of physical reasoning before organizing them into three general technical approaches and propose six categorical tasks of the field. Finally, we highlight the challenges of the current field and present some future research directions.
\end{abstract}

\section{Introduction} 

\begin{figure}[thb]

  \centering
  \centerline{\includegraphics[width=\linewidth]{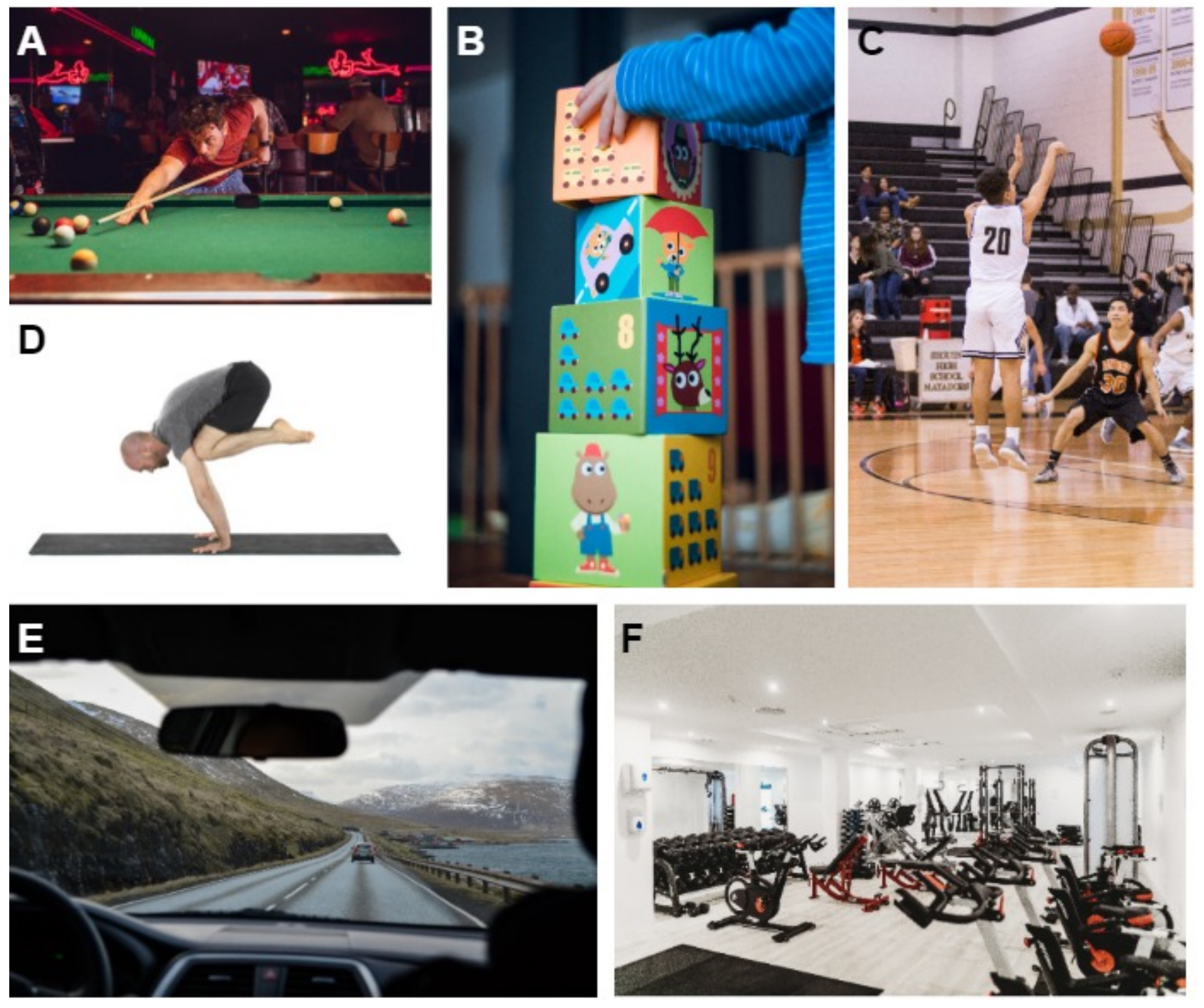}}
  \caption{Examples of everyday scenes that requires us to employ intuitive physics. (A) Predicting the trajectories of the billiard balls. (B) Balancing the stacking blocks. (C) Shooting the basketball with a parabola trajectory. (D) Balancing one's body during yoga. (E) Estimating the velocity of the car travelling ahead. (F) A gym with equipment that exhibits various degree of physical properties.  }
  
  
\label{fig:first}

\end{figure}

Humans have demonstrated the innate ability to approximate predictions of their surrounding interactions and physical environment even without any formal education in physics \cite{article} as shown in the examples from Figure \ref{fig:first}. In fact, research in developmental psychology shows that infants as young as two and half months can understand fundamental physics \cite{carey2000origin,baillargeon2004infants} and by three months old, they can detect violations of physical principles of persistence, continuity and solidity \cite{leslie1984spatiotemporal}. They achieve this separately from acquisition of semantic knowledge, language, and sensorimotor skills. This cognitive capability of humans is commonly termed \textbf{intuitive physics}, and widely used by researchers in multiple disciplines like cognitive science, neuroscience and computer science. 

For several decades, the cognitive science perspectives on intuitive physics have been shaped by the question of how humans acquire and deploy this cognitive capability. Notwithstanding various debates on how intuitive physics works, several conventional and widely plausible ideas have been proposed by cognitive scientists. They include heuristic or rule-based models \cite{gilden1994heuristic,runeson2000visual,sanborn2013reconciling}, probabilistic mental simulation \cite{hegarty2004mechanical,bates2015humans}, and the cognitive intuitive physics engine (IPE) \cite{battaglia2013simulation,ullman2017mind}, each with its own merits. With recent advancements of computing technology and motivations to create machines that can learn and think like humans \cite{lake2017building}, the modern field of intuitive physics has been reinvigorated by new techniques in artificial intelligence (AI).

This paper aims to provide a comprehensive survey of intuitive physics for machine cognition, covering the recent advancements in computer vision, deep learning, and AI deployed to model human-level intuitive physics capability for various physical reasoning tasks. This survey will first examine intuitive physics in machine cognition by categorizing existing works into three facets of physical reasoning, namely \emph{prediction, inference and causal reasoning}. Then, the paper will categorize the space of physical reasoning tasks into six categorical tasks, and further review them via their three general technical approaches (\textbf{inverse rendering, inverse physics, and inverse dynamics}) as shown in Table \ref{tab:tasks}. Lastly, the paper will then conclude with some of the open problems, challenges and future trends in intuitive physics for machine cognition.

There is a larger body of work that could constitute as intuitive physics for machine cognition. However, this survey is scoped specifically to the facets, approaches and tasks defined in later sections. For instance, physical reasoning with 3D and 2.5D data representations \cite{zheng2013beyond,jia20143d,du2018learning}, or on the non-rigid body (e.g. fluid, particle, and soft-body) \cite{li2018learning,mrowca2018flexible,li2020visual}, and learning physics via action-task planning \cite{song2018inferring,bakhtin2019phyre,xu2020learning} will not be covered in this paper.    

\begin{figure*}[thb]

  \centering
  \centerline{\includegraphics[width=16.5cm]{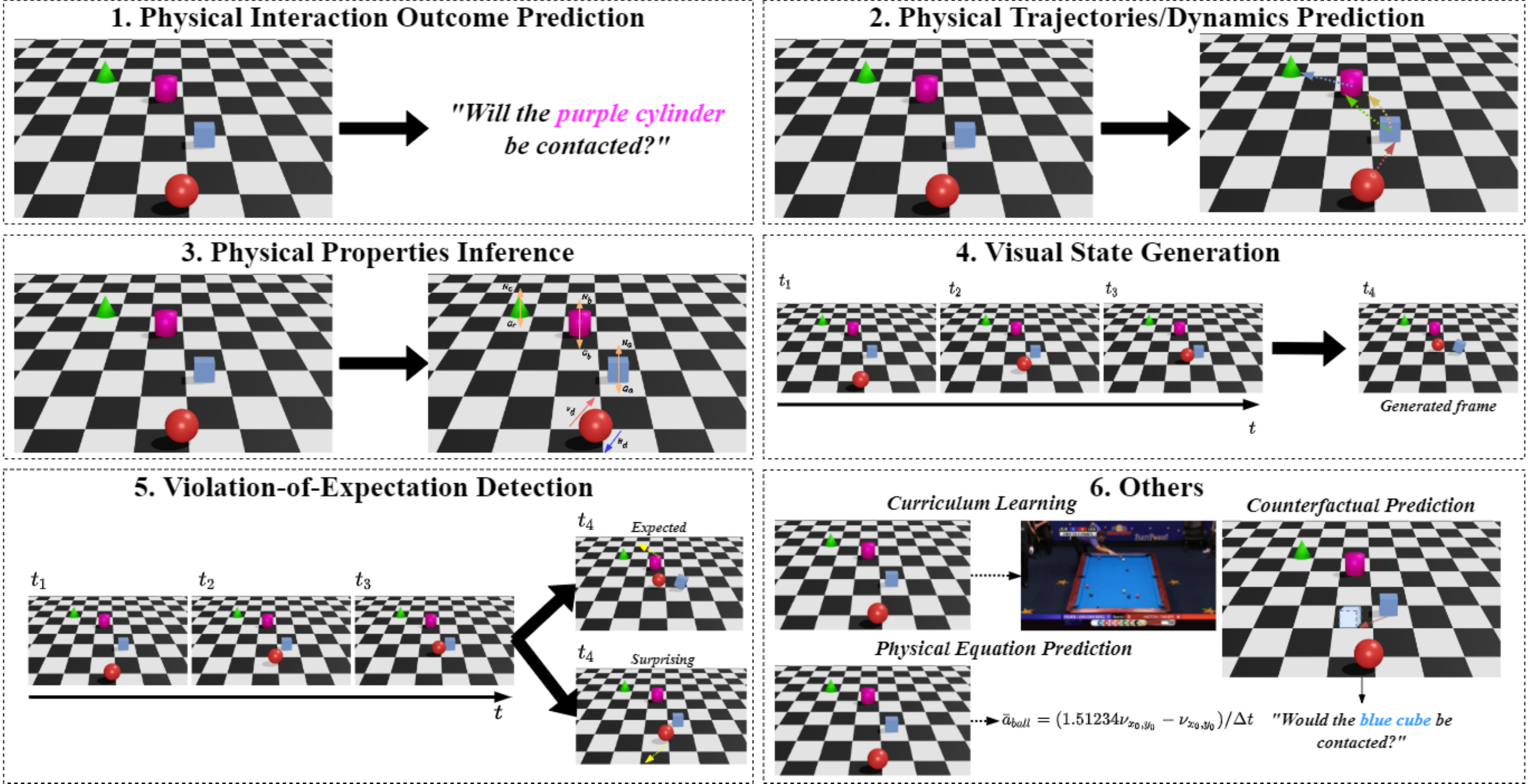}}
  \caption{Summary of the six physical reasoning tasks. (1) \textbf{PIO}, to  predict  the  different  states  or  outcome  of  physical  interactions  (e.g.,  \emph{if  objects  within  the  dynamic  scene  is stable, contained, or contacted}). (2) \textbf{PTD}, to predict the possible physical trajectories given only a few dynamics scenes. (3) \textbf{PPI}, to infer both the observed(e.g., \emph{size, color, and shape}) and latent physical properties (e.g., \emph{mass, friction, velocity, and displacement}) of the objects within the dynamic scenes. (4) \textbf{VSG}, to generate the unseen future frames of a long roll-out  sequence given  only  the  initial  few  dynamics  frames. (5) \textbf{VoE}, to classify if there  is  any  violation-of-expectation  in  a  given  dynamics scenes. (6) \textbf{Others}, other intuitive physics-inspired AI tasks such as curriculum learning, counterfactual prediction, physical equations prediction etc.}
  
  
\label{fig:connection}

\end{figure*}

\section{Background}
\subsection{Motivation}
A primary purpose of intuitive physics is to allow us to plan effective actions on the world. We have a goal in mind for a physical outcome we want to achieve, and intuitive physics allows us to assess the consequences of possible actions in order to select those that will achieve our goals. However, intuitive physics can be approximate, but it shouldn't be fundamentally wrong. But yet, humans tend to develop various misconceptions \cite{caramazza1981naive,mccloskey1983intuitive} in their physical judgement (e.g., the Aristotelian prediction). The main goal of intuitive physics in human cognition is our ability to rely on intuition and build upon the interactions with our surroundings and make adequate physical reasoning of the observed events. Prior work in cognitive science literature emphasizes the importance of intuitive physics as forms of high-level reasoning capabilities rooted in developing intelligence systems. Therefore, intuitive physics understanding is vital for AI to have a general understanding of physical scenes and ensuring the safety of embodied AI \cite{duan2021survey} systems deployed into the real-world \cite{duchaine2009safe,zheng2015scene}. Furthermore, there is a diverse amount of work in intuitive physics for machine cognition, however with a scarcity of comprehensive survey papers \cite{kubricht2017intuitive} on the field from a machine learning perspective. 

\subsection{Survey Organization}

The paper will look into current efforts in physical reasoning via a hierarchical structure by first categorizing them into three facets of physical reasoning: \emph{prediction, inference} and \emph{causal reasoning} \cite{smith_sanborn_battaglia_gerstenberg_ullman_tenenbaum}. Following that, the paper groups existing efforts for intuitive physics for machine cognition by three general technical approaches for physical reasoning tasks: \textbf{inverse rendering}, which uses a single image to extrapolate and learn useful features, \textbf{inverse physics}, which uses latent physical representation (e.g., object relation graph, physical properties, and others) to perform physical reasoning, and lastly \textbf{inverse dynamics}, which uses sequential roll-out frames as a representation of object dynamics in physical events for learning. The work under these three general approaches is further categorised based on six physical reasoning tasks widely used to evaluate intuitive physics for machine cognition. The six physical reasoning tasks are Physical Interaction Outcome prediction (PIO), Physical Trajectories/Dynamics prediction (PTD), Physical Properties Inference (PPI), Visual State Generation (VSG), Violation-of-Expectation detection (VoE), and other intuitive physics-inspired AI tasks (Others). The details of these six physical reasoning tasks are illustrated in Figure \ref{fig:connection}. The paper evaluates all existing work by their approaches, physical reasoning tasks and evaluation metrics in Section \ref{sec:birds} and Table \ref{tab:tasks}. Finally, the paper will discusses the current challenges of the field and proposes some open questions in Section \ref{sec:five}.

With a few exceptions (e.g. generative task, counterfactual prediction, or predicting physical equations), the majority of physical reasoning tasks focus on the machine learning task of classification. This is due to the broad nature of humans' ability to convey the output of our intuitive physics model through a form of classifying the various possibilities of any physical interaction and thus provide reasoning through classification.

\section{Facets of Physical Reasoning}
\label{sec:birds}
\subsection{Prediction} 
The goal of \emph{prediction} in physical reasoning, from a cognitive standpoint, is to deploy a forward intuitive physics model for physical interactions before querying on the simulated outcomes and making judgments about the possible future states of the interactions. The goal of \emph{prediction} for machine cognition focuses on a simple notion, which is taking in visual inputs and performing physical reasoning of the queried scenarios. All work cited in this section tackled this facet via deep learning. The deep neural networks (DNN) implemented generally mapped the visual inputs to the various physical prediction outputs (e.g., the outcome of the physical interactions, physical trajectories, inferred physical properties of objects and generating future frames).



\textbf{PIO}, \textbf{PTD}, \textbf{PPI}, and \textbf{VSG} predictions are the common physical reasoning tasks that fall under the facets of \emph{prediction}. Most of the work under \emph{prediction} employ either an inverse rendering or inverse dynamics approach, with one exception \cite{Mottaghi2016WhatHI}.  They focused on using conventional convolutional neural network (CNN) frameworks such as InceptionNet \cite{szegedy2015going}, AlexNet \cite{iandola2016squeezenet}, and ResNet \cite{he2016deep}
as their backbone for learning to map the input pixels into low-level features that are later used to predict the various physical reasoning tasks. As a result, many of these works came at the beginning of deep learning era.  Consequently, all of these works \cite{mottaghi2016newtonian,li2016fall,lerer2016learning,Mottaghi2016WhatHI,stack,janner2018reasoning,wat,wu2017learning,Ehrhardt2019UnsupervisedIP,duan2021pip} fall under the facet of \emph{prediction}. Within the facets of physical reasoning, the falling-tower test is one of the most common tests for evaluating a model's ability in physical reasoning. Hence, all of these works \cite{li2016fall,Mottaghi2016WhatHI,lerer2016learning,stack,janner2018reasoning,duan2021pip} focus on predicting the outcome of stability for the given dynamics scenarios. Besides just predicting the outcome of stability, Lerer \emph{et al.}~\shortcite{lerer2016learning} even predicts the trajectories of the stacking blocks via segmentation masks, while Groth \emph{et al.}~\shortcite{stack} predicts the point of instability through generating a heat-map. Duan \emph{et al.}~\shortcite{duan2021pip} predicts the outcome of physical interaction by mimic the "noisy" framework in human physical prediction with generative models, and using attention to focus on salient moments of physical interactions. 

Another common test for physical reasoning is for the model to make judgments of the physical properties for object interactions in Newtonian motion (e.g. Spring, Gravity, Billiards, Magnetic billiards, drift). As a result of this test for intuitive physics, works such as Mottaghi \emph{et al.}~\shortcite{mottaghi2016newtonian} and Watters \emph{et al.}~\shortcite{wat} generate a set of synthetic 2D video dataset using a physics engine for the models' training. Their models would focus on first learning to infer the physical properties from a given visual input using DNN and use the derived physical properties to reason and generate the potential physical trajectories or dynamics of the objects within the scene. The work in Watters \emph{et al.}~\shortcite{wat} even employs a physics engine to reverse the learned latent state back into generated frames forward in time. On the other hand, works such as Wu \emph{et al.}~\shortcite{wu2017learning} and Ehrhardt \emph{et al.}~\shortcite{Ehrhardt2019UnsupervisedIP} use this test to evaluate their model's intuitive physics ability; however, they train their models using real-world video datasets collected on these various physical interactions. Wu \emph{et al.}~\shortcite{wu2017learning} would infer object properties from visual inputs and, using a game engine, simulate these obtained physical properties back into an image. In contrast, Ehrhardt \emph{et al.}~\shortcite{Ehrhardt2019UnsupervisedIP} uses a meta-learning approach to learn from the past dynamics scenes and uses that to optimise the learning process of predicting new trajectories. While inferring physical properties is a step of their approaches, we consider these works to fall under the facet of prediction as their main goal in the tasks is predictive in nature.

\begin{table*}[!t]

\caption{\label{tasks} Summary of the work for intuitive physics for machine cognition. Physical Reasoning Tasks: Physical Interaction Outcome prediction (PIO), Physical Trajectories/Dynamics prediction (PTD), Physical Properties Inference (PPI), Visual State Generation (VSG), Violation-of-Expectation detection (VoE), other intuitive physics-inspired AI tasks (Others). The evaluation metric: F1 score ($F_1$), modified hausdorff distance (MHD), prediction accuracy ($Acc_\%$), intersection over union (IoU), log likelihood ($Log_L$), mean squared error (MSE), prediction variable error ($Error$), mean euclidean prediction error (\emph{MEPE}), inverse normalized loss ($IN_L$), coefficient of determination
 ($R^2$), structural similarity index (SSI), KL divergence (KD), mean reciprocal rank (MRR) and error rate (1-AUC).}
\centering
\resizebox{\linewidth}{!}{
\begin{tabular}{|c|c|c|c|c|c|}
\hline
Facets of Physical Reasoning & Method/Category & Publication & Year & Physical Reasoning Tasks & Evaluation Metrics \\
\hline\hline
Prediction & Inverse Rendering &  \cite{mottaghi2016newtonian} & 2016 & PPI & $F_1$, MHD \\
\cline{3-6} & & \cite{li2016fall} & 2016 & PPI & $Acc_\%$, $R^2$ \\
\cline{3-6} & & \cite{lerer2016learning} & 2016 & PIO, PTD &$Acc_\%$, IoU, $Log_L$ \\
\cline{3-6} & & \cite{stack} & 2018 & PIO, PPI & $Acc_\%$ \\
\cline{3-6} & & \cite{janner2018reasoning} & 2018 & VSG & $Acc_\%$, MSE \\

\cline{2-6} & Inverse Physics & \cite{Mottaghi2016WhatHI} & 2016 & PTD& $Acc_\%$\\
\cline{3-6} & &\cite{Ehrhardt2019UnsupervisedIP} & 2019 & PTD& MSE, $Error$\\

\cline{2-6} & Inverse Dynamics & \cite{wat} & 2017 & PTD, VSG& $IN_L$, \emph{MEPE} \\
\cline{3-6} & & \cite{wu2017learning} & 2017 & PPI & MSE, MAE \\
\cline{3-6} & & \cite{duan2021pip} & 2021 & PIO, VSG& $Acc_\%$, \emph{PSNR} \\
\hline

Inference & Inverse Rendering &  \cite{wu2015galileo} & 2015 & PIO, PPI& $Acc_\%$, MSE \\
\cline{3-6} & & \cite{Wu2016Physics1L} & 2016 & PIO, PTD& MSE, $R^2$ \\

\cline{2-6} & Inverse Physics & \cite{Battaglia2016InteractionNF} & 2016 & PPI, PTD& MSE\\
\cline{3-6} & & \cite{chang2016compositional} & 2016 & PPI& MSE, $Acc_\%$ \\
\cline{3-6} & & \cite{Huang2018PerceivingPE} & 2018 & Others& $R^2$, \emph{MEPE} \\
\cline{3-6} & & \cite{xu2021bayesian} & 2021 & PPI& IoU, MSE\\

\cline{2-6} & Inverse Dynamics & \cite{Fragkiadaki2016LearningVP} & 2016 & PPI, PTD& $Error$, $Acc_\%$  \\
\cline{3-6} & & \cite{ye2018interpretable} & 2018 & PIO, PPI& $Error$ \\
\cline{3-6} & & \cite{zheng2018unsupervised} & 2018 & PPI, VSG& $R^2$, \emph{MEPE}\\
\cline{3-6} & & \cite{de2018end} & 2018 & PPI, VSG& MSE \\
\cline{3-6} & & \cite{jaques2019physics} & 2019 & PPI, VSG& SSI \\
\cline{3-6} & & \cite{kandukuri2022physical} & 2022 & PPI& $Error$\\
\hline

Causal Reasoning & Inverse Rendering &  \cite{piloto2018probing} & 2018 & VoE& \emph{KD} \\
\cline{3-6} & & \cite{smith2019modeling} & 2019 & VoE& $Acc_\%$, RMSE  \\

\cline{2-6} & Inverse Physics & \cite{Baradel_2020_ICLR} & 2020 & VSG, Others & MSE\\
\cline{3-6} & & \cite{li2020causal} & 2020 & VSG, Others& MRR, $F_1$ \\
\cline{3-6} & & \cite{dasgupta2021benchmark} & 2021 & VoE& $Acc_\%$ \\

\cline{2-6} & Inverse Dynamics & \cite{riochet2018intphys} & 2018 & VoE& 1-AUC \\
\cline{3-6} & & \cite{ates2020craft} & 2020 & PIO, Others & $Acc_\%$ \\
\cline{3-6} & & \cite{CLEVRER2020ICLR} & 2020 & PIO, Others& $Acc_\%$ \\
\hline

\end{tabular}}
\label{tab:tasks}
\end{table*}


\subsection{Inference} 


The aim of inference in physical reasoning is to inductively represent the physical properties of a physical interaction. Consequently, most of the papers listed under the facet of \emph{inference} tackle the physical reasoning task of \textbf{PPI} (with the exception of Wu \emph{et al.}~\shortcite{Wu2016Physics1L} and Huang \emph{et al.}~\shortcite{Huang2018PerceivingPE}). A common theme of papers under this facet is that they often have an additional component of a prediction task. Although one may construe such papers to also fall under the facet of \emph{prediction}, we label these papers under \emph{inference} as they form the best representative work of inference in intuitive physics and the inference stage forms the main and crucial back-end step of the models described in them. When defining physical properties, researchers generally consider two components. 

The first component refers to extrinsic/observable physical properties (e.g. speed, size, shape, position), which are generally directly determined with little difficulty by humans. For instance, Huang \emph{et al.}~\shortcite{Huang2018PerceivingPE} employs a two-stage Faster-RCNN model to infer the position and velocity of physical scenarios as a necessary step to learn governing physics equations via symbolic regression. However, the inference of physical properties revealed by surface textures are far easier to estimate than latent properties (e.g. mass, density, friction, coefficient of restitution, spring constant) that must be inferred through the observation of physical interactions. These latent properties cannot be directly observed and often requires humans to observe the outcome of an interaction before refining their judgements of such properties. This added challenge explains why the remainder of papers under this facet attempt to infer latent properties.

The majority of \emph{inference} papers attempt to infer latent properties from physical interactions of bouncing balls, collisions, spring oscillation and an object moving on a ramp or level surface. This is because the outcome of these interactions are sensitive to latent properties like mass, coefficient of restitution (affecting the speed after a collision), friction (affecting an object's ability to slide on a surface) and spring constant (affecting an object-spring system oscillation). In Wu \emph{et al.}~\shortcite{wu2015galileo}, a generative 3D physics engine coupled with an object tracking algorithm was used to infer the mass, position, shape and friction of an object sliding down a ramp. A markov chain monte-carlo approach was used to infer the likelihood of these physical properties with a real world dataset. Inspired by this `analysis-by-synthesis' approach, Jaques \emph{et al.}~\shortcite{jaques2019physics} implemented an unsupervised deep learning encoder-decoder inverse graphics framework to infer the spring constant, gravity and mass of a 2D synthetic ball-spring system and 3-body gravitational system. Contrary to this inverse rendering approach, Wu \emph{et al.}~\shortcite{Wu2016Physics1L} learns directly from visual inputs by first inferring observable physical properties and using them as supervision to infer latent physical properties. They also contributed a real-world dataset, Physics 101, which challenges models to infer these physical properties in various physical scenarios (e.g. object on a ramp, object floating/sinking in liquid).

One limitation thus far is that these approaches do not generalize to any number of objects. Hence, Battaglia \emph{et al.}~\shortcite{Battaglia2016InteractionNF} introduced the interaction network, a learnable graph model which can reason about the interactions with any number of objects with given object and relation based representations. This network allowed for the inference of the potential energy in 2D synthetic N-body collision interactions. The interaction network was adopted by Zheng \emph{et al.}~\shortcite{zheng2018unsupervised} as a unit for a recurrent neural network (RNN) to infer the coefficient of restitution and mass of elastic and inelastic collisions. The interaction network was also adopted by Chang \emph{et al.}~\shortcite{chang2016compositional} who combined it with pairwise factorization, context selection and function composition to form the Neural Physics Engine (NPE), which can infer discrete mass values of a 2D synthetic dataset of bouncing balls.

The need to disentangle the visual properties of an object from its dynamics for the sake of generalizability was highlighted by Chang \emph{et al.}~\shortcite{chang2016compositional} as it is possible for objects to differ visually and have the same dynamics. Another form disentanglement was deemed necessary by Ye \emph{et al.}~\shortcite{ye2018interpretable}, who underscored that it is not possible to determine the exact values of mass and friction together as they are highly dependent quantities. Therefore, in their encoder-decoder deep learning approach for their 3D (in 2D video) collision dataset, they represented the mass and friction in the latent space of the bottleneck layer and staggered the training such that the model trained to infer each quantity separately.

Another approach to the facet of \emph{inference} is the use of a differentiable physics engine. de Avila Belbute-Peres \emph{et al.}~\shortcite{de2018end} proposed a 2D differentiable physics engine that is defined with a linear complementarity problem (LCP). To test their physics engine, they illustrated its capability in inferring the unknown masses of bouncing billiard balls. The approach in de Avila Belbute-Peres \emph{et al.}~\shortcite{de2018end} was capable of fast inference via an analytical solution and had higher sample efficiency, unlike the earlier mentioned data-driven approaches. Kandukuri \emph{et al.}~\shortcite{kandukuri2022physical} also used a differentiable physics engine in their model as the first step in inferring the mass and friction of various physical scenarios (block on a flat/inclined plate and block collision). Finally, Xu \emph{et al.}~\shortcite{xu2021bayesian} also addressed the issue of sample efficiently by introducing the Bayesian Symbolic Physics (BSP) model, a probabilistic learning approach that can infer the mass and friction of N-body and bouncing ball interactions with significantly fewer (10$\times$) samples.


\subsection{Causal Reasoning} 

The facet of causal reasoning in the context of physical reasoning can be defined via two paradigms: counterfactual reasoning (\textbf{Others}) and \textbf{VoE}. The counterfactual reasoning paradigm is inspired by the theory that humans reason about causal events (considering events/objects A, B and C; does A cause B to lead to C?) by mentally modelling the scenario if the absence of the proposed causal object/event would still lead to the same outcome (if A did not exist, would B still lead to C?). If the mentally simulating that in the absence of A, B will not lead to C, an agent may conclude that there is a causal link between A and B. 

As counterfactual physical reasoning only gained interest by the AI community recently, there are only a handful of related works, most of which are at the stage of proposing novel counterfactual datasets. CLEVRER \cite{CLEVRER2020ICLR} is a Visual Question Answering (VQA) dataset that was the first to have a counterfactual task for physical reasoning. The counterfactual segment of CLEVRER would ask questions like "without A, which event will not happen?", and the agent would be provided with a video with 2D objects and a few event options. Their proposed Neuro-Symbolic Dynamic Reasoning model that contained a LSTM-based question parser, Mask R-CNN video parser and a graph-based dynamics learning model set the highest benchmark performance for the counterfactual task. 

Instead of VQA, Baradel \emph{et al.}~\shortcite{Baradel_2020_ICLR} took a different approach of designing counterfactual tasks and proposed the 3D CoPhy dataset. The CoPhy dataset provides an original stream of images, which can reveal information of the confounder dynamics of the scene, after which the initial state is changed via a do-operator and the agent must visually generate the outcome. Their Graph Convolutional Network-based CoPhyNet benchmark was set in their work. Li \emph{et al.}~\shortcite{li2020causal} tackled the CoPhy benchmark by proposing the graph and deep learning-based Causal World Model that learns unsupervised relationships between the original and alternative outcomes by estimating latent confounding variables. However, they did not provide a direct comparison with CoPhyNEt. Finally, Ates \emph{et al.}~\shortcite{ates2020craft} created a purely 2D VQA-based counterfactual dataset that expanded on CLEVRER by providing more complex concepts of `cause, enable and prevent'.

The second paradigm under causal reasoning, \textbf{VoE}, comes from the idea that human infants form expectations of physical events which determines their knowledge of causal links for transformations in physical interactions \cite{bullock1982development}. Hence, they use these causal links to determine how surprised they are when presented with a plausible or implausible scene. Like counterfactual physical reasoning, \textbf{VoE} papers in physical reasoning are comparatively new and all propose their own datasets and approaches. Riochet \emph{et al.}~\shortcite{riochet2018intphys} first proposed the IntPhys dataset that provided 3D (in 2D) scenes of possible and impossible events of `object permanence', ‘shape constancy’ and ‘continuity’. The goal in \textbf{VoE} is to train an agent to recognise the expected video as less `surprising' than the surprising version. Their convolutional autoencoder and generative adversarial network models performed poorly in comparison with their adult human trials. Piloto \emph{et al.}~\shortcite{piloto2018probing} also introduced a dataset that showcased additional events of `solidity' and `containment', using a variational autoencoder approach to establish a benchmark. 

Instead of a purely deep learning approach, Smith \emph{et al.}~\shortcite{smith2019modeling} used probabilistic simulation along with approximate derendering and particle filtering for the \textbf{VoE} task on their own dataset. The model was named ADEPT and performed with high accuracy and even replicated human judgements ‘how, when and what’ traits of surprising scenes. Finally, Dasgupta \emph{et al.}~\shortcite{dasgupta2021benchmark} proposed a heuristic-based dataset with additional events in support and collision that had augmented metadata of ground-truth features and rules of the physical interaction, representing intermediate stages of reasoning. They showed how a model could potentially leverage on these heuristics to learn with higher accuracy and learn the universal causal relationships in physical reasoning.





\section{Challenges and Open Questions}

While the earlier sections show that researchers have worked significantly on intuitive physics for machine cognition, we recognise that there still exist multiple challenges in the field. 

\textbf{Unified evaluation}. We find that there is no agreed upon unified approach to evaluating systems of intuitive physics. Being a broad topic without a `cookie-cutter' definition, researchers have explored multiple tasks and approaches that one may arguably constitute as `intuitive physics'. Although researchers have attempted to define their own method of testing intuitive physics, the unification of such methods is lacking, but crucial to creating reliable and verifiable intuitive physics systems. One example is the dataset task that is used for testing models. For instance, many researchers have used the scenario of bouncing balls \cite{Battaglia2016InteractionNF,chang2016compositional,de2018end,xu2021bayesian} for the task of inference, but they created their own version of the challenge and often different metrics for evaluation (Mean Squared Error \cite{Battaglia2016InteractionNF,de2018end} and Accuracy \cite{chang2016compositional}). We encourage researchers to work with one version of the task coupled with a standardized metric, so that models by different researchers can be used for direct comparison. If feasible, an explainable metric is preferred like accuracy. One step in this direction may be to create a dataset with a large suite of all challenges widely used by researchers. The closest example is Physion \cite{bear2021physion}, which showcases a suite of different tasks on \textbf{PIO}. However, such datasets are still needed for other tasks in Figure~\ref{tasks}.

\textbf{Complex scenarios}. Another challenge in intuitive physics for machine cognition is deploying models in more complex and realistic settings. We define complexity through 3 means. One approach is to focus on intuitive physics using real world datasets. Most datasets (with the exceptions of \cite{wu2015galileo,Wu2016Physics1L,wu2017learning,ehrhardt2019unsupervised}) are run on basic and synthetic 2D or 3D physics engine simulations replicating trivial interactions. These synthetic datasets generally do not contain noise that would be found in real-world vision. The second means of added complexity would be to increase the compositionality of interactions in physical reasoning. Datasets from \cite{dasgupta2021benchmark,smith2019modeling,bear2021physion} generally split the events of the interactions into separate videos. For instance, the distinct events of `barrier' and `containment' will never be in the same video for the datasets in Dasgupta \emph{et al.}~\shortcite{dasgupta2021benchmark} and Smith \emph{et al.}~\shortcite{smith2019modeling}. More complex datasets that mix different events of physical interactions pose an additional challenge to reasoning systems as they need to parse the scene's events and reason about them separately. The third means of added complexity is the use of intuitive physical reasoning for embodied tasks \cite{duan2021survey}. Instead of simply evaluating on videos of static physical interactions, the embodiment of agents in virtual environments may provide more complex possibilities of deploying intuitive physical reasoning modules for specific tasks. For instance, inference of mass and friction can be useful in the task of pushing an object to a target location.

\textbf{Real-world utility}. When considering such potential applications of intuitive physical reasoning systems, one open question we find interesting is \textbf{``Specifically, how can intuitive physics systems be deployed in real world applications?"}. Other than allowing machines to learn and think more like humans \cite{lake2017building}, researchers have mentioned the use of intuitive physical reasoning in safe AI systems. There is potential for intuitive physics to be useful for safety applications, but we have not found detailed expositions about achieving this. For instance, how will intuitive physical reasoning systems help autonomous vehicles make more refined decisions by reasoning about interactions in their surroundings? How can physical reasoning modules help robotic manipulators interact with new objects and efficiently learn dynamical parameters? While a plethora of specific applications of intuitive physics exist, these should be discussed more widely and exhaustively. This would not only highlight the importance of intuitive physics research in machine cognition, but it would also encourage researchers to work towards real-world systems that leverage the research.

\textbf{Generalizability}. As mentioned in Lake \emph{et al.}~\shortcite{lake2017building}, one test of a universal physical reasoning system is creating a general-purpose physical simulator that can physically reason in all possible scenarios. This leads us to ask \textbf{``How can we create an integrated intuitive physical reasoning system that learns from the context of the scenario?"}. Humans often misjudge their sense of intuitive physics when the context of the interaction is not known \cite{kubricht2017intuitive}. It would stand to reason that machines too need the context of the physical interaction before making accurate judgments in a complex scenario. A general-purpose intuitive physical reasoning system would need to be able to learn the context of the situation and decide which facet of reasoning it should invoke (\emph{prediction, inference, causal}) and the type of task to tackle. We recognise that this task is extremely challenging, but we hope that researchers recognise the potential of context-driven physical reasoning tasks and systems.

In summary, the challenges and open questions will assist researchers new to the field in converging on several existing and unsolved problems, ranging from dataset development to the definition of new field-specific assessment measures. Table \ref{tab:tasks} further provides an overview of the field and the categorization of the different facets of physical reasoning, methods, and tasks. Additionally, it provides an overview of current evaluation metrics employed in many parts of physical reasoning. This can assist researchers unfamiliar with the field to select pertinent work for their desired task.

\label{sec:five}
\section{Conclusion}

To advance towards human-like learning, the integration of intuitive physics and deep learning is crucial \cite{lake2017building}. We reviewed a range of deep learning papers at the intersection of intuitive physics and AI. These papers were categorised at three levels: facets of intuitive physics, technical approach taken and physical reasoning tasks. While some papers could have multiple labels per categorization, they were organized based on the label most crucial to the deployment of the proposed model. These categorizations give structure to an otherwise amorphous and growing field, while also allowing researchers to swiftly spot gaps in certain areas. Overall, this survey may be used as a starting point in understanding how researchers define intuitive physics for machine cognition, along with challenges and future directions.

\bibliographystyle{named}
\bibliography{ijcai22}

\begin{thebibliography}{}

\bibitem[\protect\citeauthoryear{Ates \bgroup \em et al.\egroup
  }{2020}]{ates2020craft}
Tayfun Ates, Muhammed~Samil Atesoglu, Cagatay Yigit, Ilker Kesen, Mert Kobas,
  Erkut Erdem, Aykut Erdem, Tilbe Goksun, and Deniz Yuret.
\newblock Craft: A benchmark for causal reasoning about forces and
  interactions.
\newblock {\em arXiv preprint arXiv:2012.04293}, 2020.

\bibitem[\protect\citeauthoryear{Baillargeon}{2004}]{baillargeon2004infants}
Ren{\'e}e Baillargeon.
\newblock Infants' physical world.
\newblock {\em Current directions in psychological science}, 13(3):89--94,
  2004.

\bibitem[\protect\citeauthoryear{Bakhtin \bgroup \em et al.\egroup
  }{2019}]{bakhtin2019phyre}
Anton Bakhtin, Laurens van~der Maaten, Justin Johnson, Laura Gustafson, and
  Ross Girshick.
\newblock Phyre: A new benchmark for physical reasoning.
\newblock {\em arXiv preprint arXiv:1908.05656}, 2019.

\bibitem[\protect\citeauthoryear{Baradel \bgroup \em et al.\egroup
  }{2020}]{Baradel_2020_ICLR}
Fabien Baradel, Natalia Neverova, Julien Mille, Greg Mori, and Christian Wolf.
\newblock Cophy: Counterfactual learning of physical dynamics.
\newblock In {\em ICLR}, 2020.

\bibitem[\protect\citeauthoryear{Bates \bgroup \em et al.\egroup
  }{2015}]{bates2015humans}
Christopher Bates, Peter~W Battaglia, Ilker Yildirim, and Joshua~B Tenenbaum.
\newblock Humans predict liquid dynamics using probabilistic simulation.
\newblock In {\em CogSci}. Citeseer, 2015.

\bibitem[\protect\citeauthoryear{Battaglia \bgroup \em et al.\egroup
  }{2013}]{battaglia2013simulation}
Peter~W Battaglia, Jessica~B Hamrick, and Joshua~B Tenenbaum.
\newblock Simulation as an engine of physical scene understanding.
\newblock {\em Proceedings of the National Academy of Sciences},
  110(45):18327--18332, 2013.

\bibitem[\protect\citeauthoryear{Battaglia \bgroup \em et al.\egroup
  }{2016}]{Battaglia2016InteractionNF}
Peter~W. Battaglia, Razvan Pascanu, Matthew Lai, Danilo~Jimenez Rezende, and
  Koray Kavukcuoglu.
\newblock Interaction networks for learning about objects, relations and
  physics.
\newblock {\em ArXiv}, abs/1612.00222, 2016.

\bibitem[\protect\citeauthoryear{Bear \bgroup \em et al.\egroup
  }{2021}]{bear2021physion}
Daniel~M Bear, Elias Wang, Damian Mrowca, Felix~J Binder, Hsiau-Yu~Fish Tung,
  RT~Pramod, Cameron Holdaway, Sirui Tao, Kevin Smith, Fan-Yun Sun, et~al.
\newblock Physion: Evaluating physical prediction from vision in humans and
  machines.
\newblock {\em arXiv preprint arXiv:2106.08261}, 2021.

\bibitem[\protect\citeauthoryear{Bullock \bgroup \em et al.\egroup
  }{1982}]{bullock1982development}
Merry Bullock, Rochel Gelman, and Ren{\'e}e Baillargeon.
\newblock The development of causal reasoning.
\newblock {\em The developmental psychology of time}, pages 209--254, 1982.

\bibitem[\protect\citeauthoryear{Caramazza \bgroup \em et al.\egroup
  }{1981}]{caramazza1981naive}
Alfonso Caramazza, Michael McCloskey, and Bert Green.
\newblock Naive beliefs in “sophisticated” subjects: Misconceptions about
  trajectories of objects.
\newblock {\em Cognition}, 9(2):117--123, 1981.

\bibitem[\protect\citeauthoryear{Carey}{2000}]{carey2000origin}
Susan Carey.
\newblock The origin of concepts.
\newblock {\em Journal of Cognition and Development}, 1(1):37--41, 2000.

\bibitem[\protect\citeauthoryear{Chang \bgroup \em et al.\egroup
  }{2016}]{chang2016compositional}
Michael~B Chang, Tomer Ullman, Antonio Torralba, and Joshua~B Tenenbaum.
\newblock A compositional object-based approach to learning physical dynamics.
\newblock {\em arXiv preprint arXiv:1612.00341}, 2016.

\bibitem[\protect\citeauthoryear{Dasgupta \bgroup \em et al.\egroup
  }{2021}]{dasgupta2021benchmark}
Arijit Dasgupta, Jiafei Duan, Marcelo~H Ang~Jr, Yi~Lin, Su-hua Wang, Ren{\'e}e
  Baillargeon, and Cheston Tan.
\newblock A benchmark for modeling violation-of-expectation in physical
  reasoning across event categories.
\newblock {\em arXiv preprint arXiv:2111.08826}, 2021.

\bibitem[\protect\citeauthoryear{de Avila Belbute-Peres \bgroup \em et
  al.\egroup }{2018}]{de2018end}
Filipe de~Avila Belbute-Peres, Kevin Smith, Kelsey Allen, Josh Tenenbaum, and
  J~Zico Kolter.
\newblock End-to-end differentiable physics for learning and control.
\newblock {\em Advances in neural information processing systems},
  31:7178--7189, 2018.

\bibitem[\protect\citeauthoryear{Du \bgroup \em et al.\egroup
  }{2018}]{du2018learning}
Yilun Du, Zhijian Liu, Hector Basevi, Ales Leonardis, Bill Freeman, Josh
  Tenenbaum, and Jiajun Wu.
\newblock Learning to exploit stability for 3d scene parsing.
\newblock In {\em NeurIPS}, pages 1733--1743, 2018.

\bibitem[\protect\citeauthoryear{Duan \bgroup \em et al.\egroup
  }{2021a}]{duan2021pip}
Jiafei Duan, Samson Yu, Soujanya Poria, Bihan Wen, and Cheston Tan.
\newblock Pip: Physical interaction prediction via mental imagery with span
  selection.
\newblock {\em arXiv preprint arXiv:2109.04683}, 2021.

\bibitem[\protect\citeauthoryear{Duan \bgroup \em et al.\egroup
  }{2021b}]{duan2021survey}
Jiafei Duan, Samson Yu, Hui~Li Tan, Hongyuan Zhu, and Cheston Tan.
\newblock A survey of embodied ai: From simulators to research tasks.
\newblock {\em arXiv preprint arXiv:2103.04918}, 2021.

\bibitem[\protect\citeauthoryear{Duchaine and
  Gosselin}{2009}]{duchaine2009safe}
Vincent Duchaine and Cl{\'e}ment Gosselin.
\newblock Safe, stable and intuitive control for physical human-robot
  interaction.
\newblock In {\em 2009 IEEE International Conference on Robotics and
  Automation}, pages 3383--3388. IEEE, 2009.

\bibitem[\protect\citeauthoryear{Ehrhardt \bgroup \em et al.\egroup
  }{2019a}]{ehrhardt2019unsupervised}
S{\'e}bastien Ehrhardt, Aron Monszpart, Niloy~J Mitra, and Andrea Vedaldi.
\newblock Unsupervised intuitive physics from past experiences.
\newblock {\em arXiv preprint arXiv:1905.10793}, 2019.

\bibitem[\protect\citeauthoryear{Ehrhardt \bgroup \em et al.\egroup
  }{2019b}]{Ehrhardt2019UnsupervisedIP}
S{\'e}bastien Ehrhardt, Aron Monszpart, Niloy~Jyoti Mitra, and Andrea Vedaldi.
\newblock Unsupervised intuitive physics from past experiences.
\newblock {\em ArXiv}, abs/1905.10793, 2019.

\bibitem[\protect\citeauthoryear{Fragkiadaki \bgroup \em et al.\egroup
  }{2016}]{Fragkiadaki2016LearningVP}
Katerina Fragkiadaki, Pulkit Agrawal, Sergey Levine, and Jitendra Malik.
\newblock Learning visual predictive models of physics for playing billiards.
\newblock {\em CoRR}, abs/1511.07404, 2016.

\bibitem[\protect\citeauthoryear{Gilden and
  Proffitt}{1994}]{gilden1994heuristic}
David~L Gilden and Dennis~R Proffitt.
\newblock Heuristic judgment of mass ratio in two-body collisions.
\newblock {\em Perception \& Psychophysics}, 56(6):708--720, 1994.

\bibitem[\protect\citeauthoryear{Groth \bgroup \em et al.\egroup
  }{2018}]{stack}
Oliver Groth, Fabian~B Fuchs, Ingmar Posner, and Andrea Vedaldi.
\newblock Shapestacks: Learning vision-based physical intuition for generalised
  object stacking.
\newblock In {\em European Conference on Computer Vision}, pages 724--739.
  Springer, 2018.

\bibitem[\protect\citeauthoryear{He \bgroup \em et al.\egroup
  }{2016}]{he2016deep}
Kaiming He, Xiangyu Zhang, Shaoqing Ren, and Jian Sun.
\newblock Deep residual learning for image recognition.
\newblock In {\em Proceedings of the IEEE conference on computer vision and
  pattern recognition}, pages 770--778, 2016.

\bibitem[\protect\citeauthoryear{Hegarty}{2004}]{hegarty2004mechanical}
Mary Hegarty.
\newblock Mechanical reasoning by mental simulation.
\newblock {\em Trends in cognitive sciences}, 8(6):280--285, 2004.

\bibitem[\protect\citeauthoryear{Huang \bgroup \em et al.\egroup
  }{2018}]{Huang2018PerceivingPE}
Siyu Huang, Zhi-Qi Cheng, Xi~Li, Xiao Wu, Zhongfei Zhang, and Alexander
  Hauptmann.
\newblock Perceiving physical equation by observing visual scenarios.
\newblock {\em ArXiv}, abs/1811.12238, 2018.

\bibitem[\protect\citeauthoryear{Iandola \bgroup \em et al.\egroup
  }{2016}]{iandola2016squeezenet}
Forrest~N Iandola, Song Han, Matthew~W Moskewicz, Khalid Ashraf, William~J
  Dally, and Kurt Keutzer.
\newblock Squeezenet: Alexnet-level accuracy with 50x fewer parameters and< 0.5
  mb model size.
\newblock {\em arXiv preprint arXiv:1602.07360}, 2016.

\bibitem[\protect\citeauthoryear{Janner \bgroup \em et al.\egroup
  }{2018}]{janner2018reasoning}
Michael Janner, Sergey Levine, William~T Freeman, Joshua~B Tenenbaum, Chelsea
  Finn, and Jiajun Wu.
\newblock Reasoning about physical interactions with object-oriented prediction
  and planning.
\newblock {\em arXiv preprint arXiv:1812.10972}, 2018.

\bibitem[\protect\citeauthoryear{Jaques \bgroup \em et al.\egroup
  }{2019}]{jaques2019physics}
Miguel Jaques, Michael Burke, and Timothy Hospedales.
\newblock Physics-as-inverse-graphics: Unsupervised physical parameter
  estimation from video.
\newblock {\em arXiv preprint arXiv:1905.11169}, 2019.

\bibitem[\protect\citeauthoryear{Jia \bgroup \em et al.\egroup
  }{2014}]{jia20143d}
Zhaoyin Jia, Andrew~C Gallagher, Ashutosh Saxena, and Tsuhan Chen.
\newblock 3d reasoning from blocks to stability.
\newblock {\em IEEE transactions on pattern analysis and machine intelligence},
  37(5):905--918, 2014.

\bibitem[\protect\citeauthoryear{Kandukuri \bgroup \em et al.\egroup
  }{2022}]{kandukuri2022physical}
Rama~Krishna Kandukuri, Jan Achterhold, Michael Moeller, and Joerg Stueckler.
\newblock Physical representation learning and parameter identification from
  video using differentiable physics.
\newblock {\em International Journal of Computer Vision}, 130(1):3--16, 2022.

\bibitem[\protect\citeauthoryear{Kubricht \bgroup \em et al.\egroup
  }{2017}]{kubricht2017intuitive}
James~R Kubricht, Keith~J Holyoak, and Hongjing Lu.
\newblock Intuitive physics: Current research and controversies.
\newblock {\em Trends in cognitive sciences}, 21(10):749--759, 2017.

\bibitem[\protect\citeauthoryear{Lake \bgroup \em et al.\egroup
  }{2017}]{lake2017building}
Brenden~M Lake, Tomer~D Ullman, Joshua~B Tenenbaum, and Samuel~J Gershman.
\newblock Building machines that learn and think like people.
\newblock {\em Behavioral and brain sciences}, 40, 2017.

\bibitem[\protect\citeauthoryear{Lerer \bgroup \em et al.\egroup
  }{2016}]{lerer2016learning}
Adam Lerer, Sam Gross, and Rob Fergus.
\newblock Learning physical intuition of block towers by example.
\newblock In {\em International conference on machine learning}, pages
  430--438. PMLR, 2016.

\bibitem[\protect\citeauthoryear{Leslie}{1984}]{leslie1984spatiotemporal}
Alan~M Leslie.
\newblock Spatiotemporal continuity and the perception of causality in infants.
\newblock {\em Perception}, 13(3):287--305, 1984.

\bibitem[\protect\citeauthoryear{Li \bgroup \em et al.\egroup
  }{2016}]{li2016fall}
Wenbin Li, Seyedmajid Azimi, Ale{\v{s}} Leonardis, and Mario Fritz.
\newblock To fall or not to fall: A visual approach to physical stability
  prediction.
\newblock {\em arXiv preprint arXiv:1604.00066}, 2016.

\bibitem[\protect\citeauthoryear{Li \bgroup \em et al.\egroup
  }{2018}]{li2018learning}
Yunzhu Li, Jiajun Wu, Russ Tedrake, Joshua~B Tenenbaum, and Antonio Torralba.
\newblock Learning particle dynamics for manipulating rigid bodies, deformable
  objects, and fluids.
\newblock {\em arXiv preprint arXiv:1810.01566}, 2018.

\bibitem[\protect\citeauthoryear{Li \bgroup \em et al.\egroup
  }{2020a}]{li2020causal}
Minne Li, Mengyue Yang, Furui Liu, Xu~Chen, Zhitang Chen, and Jun Wang.
\newblock Causal world models by unsupervised deconfounding of physical
  dynamics.
\newblock {\em arXiv preprint arXiv:2012.14228}, 2020.

\bibitem[\protect\citeauthoryear{Li \bgroup \em et al.\egroup
  }{2020b}]{li2020visual}
Yunzhu Li, Toru Lin, Kexin Yi, Daniel Bear, Daniel Yamins, Jiajun Wu, Joshua
  Tenenbaum, and Antonio Torralba.
\newblock Visual grounding of learned physical models.
\newblock In {\em International conference on machine learning}, pages
  5927--5936. PMLR, 2020.

\bibitem[\protect\citeauthoryear{McCloskey \bgroup \em et al.\egroup
  }{1983a}]{article}
Michael McCloskey, Allyson Washburn, and Linda Felch.
\newblock Intuitive physics: The straight-down belief and its origin.
\newblock {\em Journal of experimental psychology. Learning, memory, and
  cognition}, 9:636--49, 11 1983.

\bibitem[\protect\citeauthoryear{McCloskey \bgroup \em et al.\egroup
  }{1983b}]{mccloskey1983intuitive}
Michael McCloskey, Allyson Washburn, and Linda Felch.
\newblock Intuitive physics: the straight-down belief and its origin.
\newblock {\em Journal of Experimental Psychology: Learning, Memory, and
  Cognition}, 9(4):636, 1983.

\bibitem[\protect\citeauthoryear{Mottaghi \bgroup \em et al.\egroup
  }{2016a}]{mottaghi2016newtonian}
Roozbeh Mottaghi, Hessam Bagherinezhad, Mohammad Rastegari, and Ali Farhadi.
\newblock Newtonian scene understanding: Unfolding the dynamics of objects in
  static images.
\newblock In {\em Proceedings of the IEEE Conference on Computer Vision and
  Pattern Recognition}, pages 3521--3529, 2016.

\bibitem[\protect\citeauthoryear{Mottaghi \bgroup \em et al.\egroup
  }{2016b}]{Mottaghi2016WhatHI}
Roozbeh Mottaghi, Mohammad Rastegari, Abhinav~Kumar Gupta, and Ali Farhadi.
\newblock "what happens if..." learning to predict the effect of forces in
  images.
\newblock {\em ArXiv}, abs/1603.05600, 2016.

\bibitem[\protect\citeauthoryear{Mrowca \bgroup \em et al.\egroup
  }{2018}]{mrowca2018flexible}
Damian Mrowca, Chengxu Zhuang, Elias Wang, Nick Haber, Li~Fei-Fei, Joshua~B
  Tenenbaum, and Daniel~LK Yamins.
\newblock Flexible neural representation for physics prediction.
\newblock {\em arXiv preprint arXiv:1806.08047}, 2018.

\bibitem[\protect\citeauthoryear{Piloto \bgroup \em et al.\egroup
  }{2018}]{piloto2018probing}
Luis Piloto, Ari Weinstein, Dhruva TB, Arun Ahuja, Mehdi Mirza, Greg Wayne,
  David Amos, Chia-chun Hung, and Matt Botvinick.
\newblock Probing physics knowledge using tools from developmental psychology.
\newblock {\em arXiv preprint arXiv:1804.01128}, 2018.

\bibitem[\protect\citeauthoryear{Riochet \bgroup \em et al.\egroup
  }{2018}]{riochet2018intphys}
Ronan Riochet, Mario~Ynocente Castro, Mathieu Bernard, Adam Lerer, Rob Fergus,
  V{\'e}ronique Izard, and Emmanuel Dupoux.
\newblock Intphys: A framework and benchmark for visual intuitive physics
  reasoning.
\newblock {\em arXiv preprint arXiv:1803.07616}, 2018.

\bibitem[\protect\citeauthoryear{Runeson \bgroup \em et al.\egroup
  }{2000}]{runeson2000visual}
Sverker Runeson, Peter Juslin, and Henrik Olsson.
\newblock Visual perception of dynamic properties: Cue heuristics versus
  direct-perceptual competence.
\newblock {\em Psychological review}, 107(3):525, 2000.

\bibitem[\protect\citeauthoryear{Sanborn \bgroup \em et al.\egroup
  }{2013}]{sanborn2013reconciling}
Adam~N Sanborn, Vikash~K Mansinghka, and Thomas~L Griffiths.
\newblock Reconciling intuitive physics and newtonian mechanics for colliding
  objects.
\newblock {\em Psychological review}, 120(2):411, 2013.

\bibitem[\protect\citeauthoryear{Smith \bgroup \em et al.\egroup
  }{2019}]{smith2019modeling}
Kevin Smith, Lingjie Mei, Shunyu Yao, Jiajun Wu, Elizabeth Spelke, Joshua
  Tenenbaum, and Tomer Ullman.
\newblock Modeling expectation violation in intuitive physics with coarse
  probabilistic object representations.
\newblock 2019.

\bibitem[\protect\citeauthoryear{Smith \bgroup \em et al.\egroup }{in
  press}]{smith_sanborn_battaglia_gerstenberg_ullman_tenenbaum}
KA~Smith, AN~Sanborn, PW~Battaglia, T~Gerstenberg, TD~Ullman, and JB~Tenenbaum.
\newblock {\em Probabilistic Models of Physical Reasoning}.
\newblock in press.

\bibitem[\protect\citeauthoryear{Song and Boularias}{2018}]{song2018inferring}
Changkyu Song and Abdeslam Boularias.
\newblock Inferring 3d shapes of unknown rigid objects in clutter through
  inverse physics reasoning.
\newblock {\em IEEE Robotics and Automation Letters}, 4(2):201--208, 2018.

\bibitem[\protect\citeauthoryear{Szegedy \bgroup \em et al.\egroup
  }{2015}]{szegedy2015going}
Christian Szegedy, Wei Liu, Yangqing Jia, Pierre Sermanet, Scott Reed, Dragomir
  Anguelov, Dumitru Erhan, Vincent Vanhoucke, and Andrew Rabinovich.
\newblock Going deeper with convolutions.
\newblock In {\em Proceedings of the IEEE conference on computer vision and
  pattern recognition}, pages 1--9, 2015.

\bibitem[\protect\citeauthoryear{Ullman \bgroup \em et al.\egroup
  }{2017}]{ullman2017mind}
Tomer~D Ullman, Elizabeth Spelke, Peter Battaglia, and Joshua~B Tenenbaum.
\newblock Mind games: Game engines as an architecture for intuitive physics.
\newblock {\em Trends in cognitive sciences}, 21(9):649--665, 2017.

\bibitem[\protect\citeauthoryear{Watters \bgroup \em et al.\egroup
  }{2017}]{wat}
Nicholas Watters, Daniel Zoran, Theophane Weber, Peter Battaglia, Razvan
  Pascanu, and Andrea Tacchetti.
\newblock Visual interaction networks: Learning a physics simulator from video.
\newblock {\em Advances in neural information processing systems},
  30:4539--4547, 2017.

\bibitem[\protect\citeauthoryear{Wu \bgroup \em et al.\egroup
  }{2015}]{wu2015galileo}
Jiajun Wu, Ilker Yildirim, Joseph~J Lim, Bill Freeman, and Josh Tenenbaum.
\newblock Galileo: Perceiving physical object properties by integrating a
  physics engine with deep learning.
\newblock {\em Advances in neural information processing systems}, 28:127--135,
  2015.

\bibitem[\protect\citeauthoryear{Wu \bgroup \em et al.\egroup
  }{2016}]{Wu2016Physics1L}
Jiajun Wu, Joseph~J. Lim, Hongyi Zhang, Joshua~B. Tenenbaum, and William~T.
  Freeman.
\newblock Physics 101: Learning physical object properties from unlabeled
  videos.
\newblock In {\em BMVC}, 2016.

\bibitem[\protect\citeauthoryear{Wu \bgroup \em et al.\egroup
  }{2017}]{wu2017learning}
Jiajun Wu, Erika Lu, Pushmeet Kohli, Bill Freeman, and Josh Tenenbaum.
\newblock Learning to see physics via visual de-animation.
\newblock {\em Advances in Neural Information Processing Systems}, 30:153--164,
  2017.

\bibitem[\protect\citeauthoryear{Xu \bgroup \em et al.\egroup
  }{2020}]{xu2020learning}
Zhenjia Xu, Zhanpeng He, Jiajun Wu, and Shuran Song.
\newblock Learning 3d dynamic scene representations for robot manipulation.
\newblock {\em arXiv preprint arXiv:2011.01968}, 2020.

\bibitem[\protect\citeauthoryear{Xu \bgroup \em et al.\egroup
  }{2021}]{xu2021bayesian}
Kai Xu, Akash Srivastava, Dan Gutfreund, Felix Sosa, Tomer Ullman, Josh
  Tenenbaum, and Charles Sutton.
\newblock A bayesian-symbolic approach to reasoning and learning in intuitive
  physics.
\newblock {\em Advances in Neural Information Processing Systems}, 34, 2021.

\bibitem[\protect\citeauthoryear{Ye \bgroup \em et al.\egroup
  }{2018}]{ye2018interpretable}
Tian Ye, Xiaolong Wang, James Davidson, and Abhinav Gupta.
\newblock Interpretable intuitive physics model.
\newblock In {\em Proceedings of the European Conference on Computer Vision
  (ECCV)}, pages 87--102, 2018.

\bibitem[\protect\citeauthoryear{Yi \bgroup \em et al.\egroup
  }{2020}]{CLEVRER2020ICLR}
Kexin Yi, Chuang Gan, Yunzhu Li, Pushmeet Kohli, Jiajun Wu, Antonio Torralba,
  and Joshua~B. Tenenbaum.
\newblock {CLEVRER:} collision events for video representation and reasoning.
\newblock In {\em ICLR}, 2020.

\bibitem[\protect\citeauthoryear{Zheng \bgroup \em et al.\egroup
  }{2013}]{zheng2013beyond}
Bo~Zheng, Yibiao Zhao, Joey~C Yu, Katsushi Ikeuchi, and Song-Chun Zhu.
\newblock Beyond point clouds: Scene understanding by reasoning geometry and
  physics.
\newblock In {\em Proceedings of the IEEE Conference on Computer Vision and
  Pattern Recognition}, pages 3127--3134, 2013.

\bibitem[\protect\citeauthoryear{Zheng \bgroup \em et al.\egroup
  }{2015}]{zheng2015scene}
Bo~Zheng, Yibiao Zhao, Joey Yu, Katsushi Ikeuchi, and Song-Chun Zhu.
\newblock Scene understanding by reasoning stability and safety.
\newblock {\em International Journal of Computer Vision}, 112(2):221--238,
  2015.

\bibitem[\protect\citeauthoryear{Zheng \bgroup \em et al.\egroup
  }{2018}]{zheng2018unsupervised}
David Zheng, Vinson Luo, Jiajun Wu, and Joshua~B Tenenbaum.
\newblock Unsupervised learning of latent physical properties using
  perception-prediction networks.
\newblock {\em arXiv preprint arXiv:1807.09244}, 2018.

\end{thebibliography}

\end{document}